\documentclass[letterpaper]{article} 
\usepackage{aaai24}  
\usepackage{times}  
\usepackage{helvet}  
\usepackage{courier}  
\usepackage[hyphens]{url}  
\usepackage{graphicx} 
\usepackage{natbib}  
\usepackage{caption} 
\usepackage{algorithm}
\usepackage{algorithmic}

\usepackage{bbding}
\usepackage{xcolor}
\usepackage{multirow}
\usepackage{pifont}
\usepackage{graphicx} 
\usepackage{float} 
\usepackage{subfigure} 
\usepackage{booktabs}
\usepackage{colortbl}
\usepackage[export]{adjustbox}
\usepackage{caption}
\usepackage{array}
\usepackage{amsmath}
\usepackage{ragged2e}
\usepackage{amssymb}
\usepackage{adjustbox}
\usepackage{color}
\usepackage{tabularray}
\usepackage{CJKutf8}

\usepackage{tabularx}
\usepackage{float}
\usepackage{pythonhighlight}

\newcommand\ourmethod{MCL-NER}
\newcommand\task{CrossNER}

\definecolor{loc}{RGB}{147,112,219}
\definecolor{per}{RGB}{240,128,128}
\definecolor{org}{RGB}{255,222,173}
\definecolor{misc}{RGB}{255,224,178}
\usepackage{newfloat}
\usepackage{listings}
\DeclareCaptionStyle{ruled}{labelfont=normalfont,labelsep=colon,strut=off} 
\floatstyle{ruled}
\newfloat{listing}{tb}{lst}{}
\floatname{listing}{Listing}
\title{MCL-NER: Cross-Lingual Named Entity Recognition via\\Multi-View Contrastive Learning}
\author{
    Ying Mo\textsuperscript{\rm 1}, 
    Jian Yang\textsuperscript{\rm 1}\equalcontrib ,
    Jiahao Liu\textsuperscript{\rm 2},
    Qifan Wang\textsuperscript{\rm 3}, 
    Ruoyu Chen\textsuperscript{\rm 4},
    Jingang Wang\textsuperscript{\rm 2}, 
    Zhoujun Li\textsuperscript{\rm 1}\equalcontrib 
}
\affiliations{

    \textsuperscript{\rm 1}State Key Lab of Software Development Environment, Beihang University, Beijing, China \\  
    \textsuperscript{\rm 2}Meituan, Beijing, China \\
    \textsuperscript{\rm 3}Meta AI, New York, United States \\
    \textsuperscript{\rm 4}Beijing Information Science and Technology University, Beijing, China \\
    \{moying, jiaya, lizj\}@buaa.edu.cn, \{liujiahao12, wangjingang02\}@meituan.com, wqfcr@fb.com, chenruoyu@bistu.edu.cn
}

\usepackage{bibentry}

\begin{document}

\maketitle
\begin{abstract}
Cross-lingual named entity recognition (\task{}) faces challenges stemming from uneven performance due to the scarcity of multilingual corpora, especially for non-English data. While prior efforts mainly focus on data-driven transfer methods, a significant aspect that has not been fully explored is aligning both semantic and token-level representations across diverse languages.
In this paper, we propose \textbf{M}ulti-view \textbf{C}ontrastive \textbf{L}earning for Cross-lingual \textbf{N}amed \textbf{E}ntity \textbf{R}ecognition (\ourmethod{}). Specifically, we reframe the \task{} task into a problem of recognizing relationships between pairs of tokens. This approach taps into the inherent contextual nuances of token-to-token connections within entities, allowing us to align representations across different languages. A multi-view contrastive learning framework is introduced to encompass semantic contrasts between source, codeswitched, and target sentences, as well as contrasts among token-to-token relations. By enforcing agreement within both semantic and relational spaces, we minimize the gap between source sentences and their counterparts of both codeswitched and target sentences. This alignment extends to the relationships between diverse tokens, enhancing the projection of entities across languages. We further augment \task{} by combining self-training with labeled source data and unlabeled target data. Our experiments on the XTREME benchmark, spanning 40 languages, demonstrate the superiority of \ourmethod{} over prior data-driven and model-based approaches. It achieves a substantial increase of nearly +2.0 $F_1$ scores across a broad spectrum and establishes itself as the new state-of-the-art performer.
\end{abstract}

\section{Introduction}
\begin{figure}[t] 
    \centering 
    \includegraphics[width=0.95\columnwidth]{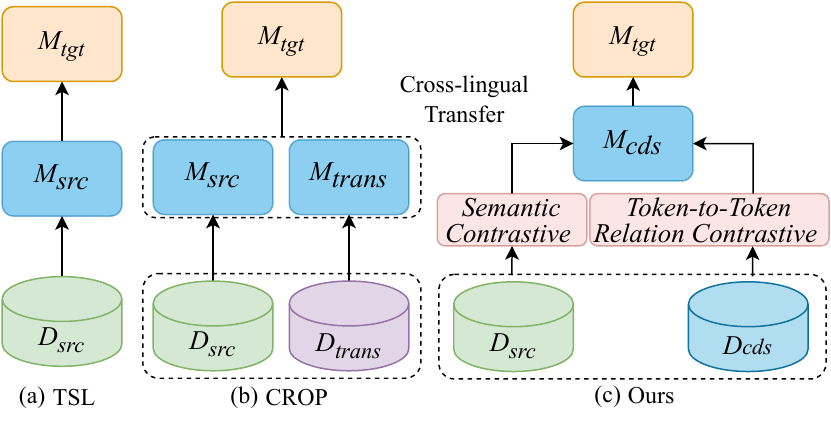} 
    \caption{Illustration of \ourmethod{} vs. existing methods TSL \cite{Multi_source_2020} and CROP \cite{CROP_Yang_2022}. $D_{src}$, $D_{trans}$ and $D_{cds}$ are the source, translated-source and code-switched data respectively. $M_{*}$ represents the trained models from the corresponding data. Our model leverages multi-view contrastive learning to bridge the gap between cross-lingual semantic and token-to-token representations.
 } 
\label{Fig.frame_1}
\end{figure}
Cross-lingual named entity recognition (\task{}) suffered from significant performance degradation in low-resource languages with limited data. 
In response to this challenge, the advent of multilingual pre-trained models \cite{bert,xlmr,alm,xmt,wang2,xlmt} has driven the development of model-based methods \cite{adversarial,BariJJ20,Multi_source_2020,wang2,wang3,wu2020enhanced}. 
These methods aim to facilitate knowledge transfer from languages with ample resources to those with fewer resources. 
Furthermore, recent studies \cite{unitrans,CROP_Yang_2022,ConNER_Zhou_2022,zhou2023improving} unify model-based and data-based transfer methods to enhance \task{}. The efficacy of these methods hinges on both the inherent cross-lingual abilities of the pre-trained models and the quality of the synthetic data produced via phrase-level and sentence-level translation.

Along the research line of levering the cross-lingual pre-trained model, previous works \cite{Mayhew_2017,Xie_2018,unitrans,wang4,Chen_2021,CROP_Yang_2022} perform phrase-level and sentence-level translation and annotate corresponding target entities. These methods can be broadly classified into two groups, as illustrated in Figure \ref{Fig.frame_1}. Within the first group, methods like TSL \cite{Multi_source_2020} generate soft labels in the target language based on the model trained using the source language. These soft labels are then utilized to facilitate training a NER model for the target language. In contrast, techniques belonging to the second group, such as CROP \cite{CROP_Yang_2022}, leverage both the source model and translation data to fine-tune the NER model for the target domain.
Despite their effectiveness, most existing studies focus primarily on aligning semantic spaces, often disregarding the crucial cross-lingual syntactic context encompassing token-to-token relationships within entities. Yet, these relations play a pivotal role in cross-lingual NER learning, given that the structure of token-to-token relationships within bilingual sentences should exhibit similar patterns. Consequently, formulating a \task{} framework capable of capturing token-level contextual nuances across diverse languages emerges as a significant challenge.

To address the above limitation, we propose a novel multi-view contrastive learning framework for cross-lingual named entity recognition in this work. We reformulate the problem of entity recognition into token-to-token relation classification. A multi-view contrastive learning with source-codeswitch semantic contrastive and token-to-token relation contrastive is employed to train the cross-lingual NER model. 
Specifically, the token-to-token relation aspect concentrates on both intra-entity and extra-entity relationships. In the context of intra-entity relations, two tokens' adjacency is scrutinized, along with determining if they signify the start-end relationship, i.e., whether they represent the initial and concluding words of the same entity. Conversely, extra-entity relations between tokens signify that they do not pertain to the same entity.
To improve the representation of entities across different languages, we introduce sentence-level semantics source-codeswitch contrastive learning. Within this framework, code-switching sentences simultaneously incorporate source and target tokens, promoting more effective cross-lingual knowledge transfer. Further, the multilingual NER model can be iteratively trained on the distilled multilingual corpora from the initial multilingual model, which makes full use of unlabeled training data to improve the performance of the \task{}. 
We conduct experiments on the XTREME benchmark covering 40 languages and then test on the CoNLL benchmark of 4 languages. Experimental results demonstrate that our method significantly outperforms most cross-lingual sequence labeling and span-based methods. Extensive probing experiments further analyze how our method can benefit the token-level cross-lingual representation by encouraging the consistency of different languages. 

Our main contributions are summarized as follows: 
\begin{itemize}
    \item We develop a multi-view contrastive learning approach with both sentence-level and token-level aligning, simultaneously enhancing the semantic and entity representation of different languages.
    \item We introduce the code-switched data together with the source data to jointly conduct the cross-lingual transfer, which effectively captures the relationships between tokens in entities, improving the \task{} performance.
    \item We conduct comprehensive experiments on two benchmarks, demonstrating competitive cross-lingual NER performance, establishing new state-of-the-art results on most of the evaluated cross-lingual transfer pairs (XTREME-40 and CoNLL).
\end{itemize}

\section{Related work}
\subsubsection{Cross-lingual NER} Cross-language named entity recognition (\task{}) achieves significant progress in recent years due to the development of pre-trained language models \cite{Ni_2017,Mayhew_2017,Xie_2018,wu-Dredze,Yu20_biaffine,hu2020xtreme,unitrans,Multi_source_2020,mulda,han-etal-2022-cross,ConNER_Zhou_2022,soft_template,CROP_Yang_2022}. 
The \task{} methods is mainly divided into two categories: model-transfer and data-transfer methods. Model-transfer methods \cite{Xie_2018} generally use language features to train a NER model on the labeled source language data and then directly uses it on the target language data. 
These features include aligned word representations \cite{Ni_2017,wu-Dredze,li2021unsupervised}, Wikifier features \cite{Mayhew_2017}, and meta-learning \cite{wu2020enhanced}. 
Data-transfer methods \cite{unitrans,Multi_source_2020,ConNER_Zhou_2022,um4,hltmt,CROP_Yang_2022,wang2023mt4crossoie,chai2024xcot} construct pseudo-labels for target language data by translating data \cite{unitrans,lvp_m3,wmt21_microsoft} from the source language typically. Further, self-training can continue to solve the lack of target data based on the existing trained \task{} model. But these methods usually use sequence tagging and fail to model the impact of token-to-token relationships in NER.
\subsubsection{Contrastive Learning} Contrastive learning is used in computer vision for image classification \cite{cl_vr_2020,cl_he2020momentum,cl_khosla2020supervised} and is now widely used in various tasks \cite{cl_chuang2020debiased,cl_giorgi2020declutr,cl_slot_hou2020fewshot,cl_gao2021simcse,cl_das2021container,cl_chen2022dictbert}. 
In NLP, \cite{cl_giorgi2020declutr,cl_gao2021simcse,cl_chen2022dictbert} propose to enhance semantic representation and a pre-trained model based on contrastive learning. 
\citet{cl_slot_hou2020fewshot} apply contrastive learning for slot filling and \citet{cl_das2021container} propose CONTaiNER for few-shot NER combining contrastive learning with Gaussian distribution. 
Contrastive learning can effectively pull the distance between positive samples and push the distance between negative samples to achieve better recognition results.

\section{Problem Formulation}
\label{sec:preliminary}
\begin{figure*}[t] 
	\centering 
	\includegraphics[width=0.9\textwidth]{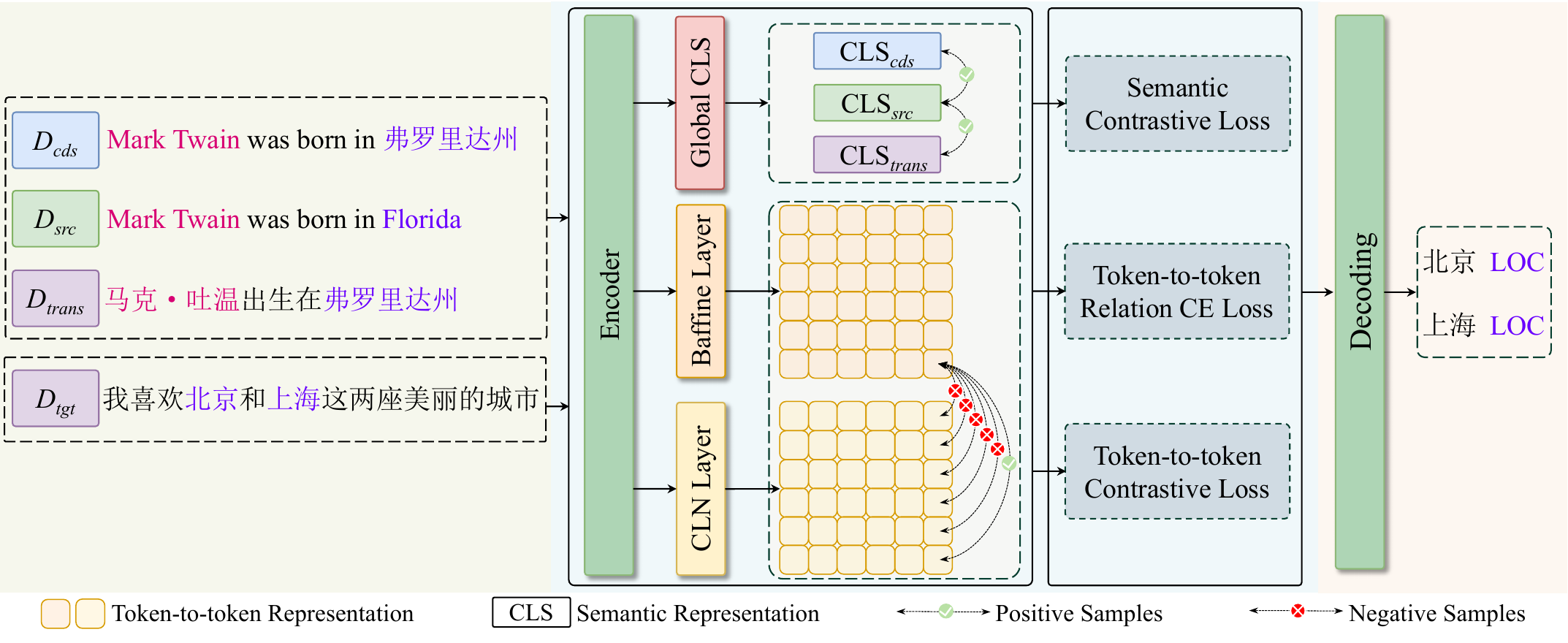} 
	\caption{Overview of the proposed \ourmethod{}. It transforms the sequence labeling problem to the classification of the token pair relation with semantic contrastive learning and token-to-token relation contrastive learning.} 
	\label{Fig.frame_2} 
\end{figure*}

\task{} aims to extract entities from target language sentences and assign them to predefined categories, without labeled data specific to the target language.
Specifically, given the target sentence $X = \{x_1,\dots,x_N\}$, each token is associated with a corresponding label $t = {t_1,\dots, t_N }$ (Tags follow the BOI schema, such as \texttt{B-LOC}, \texttt{I-LOC}, \texttt{O}), turning this task into a sequence labeling task \cite{unitrans,CROP_Yang_2022}. 
Building on the insights from prior research \cite{UniRel,PL-Marker_IE_1,OneRel,bs,w2ner,yingmo}, our approach tackles the \task{} problem by establishing relationships between all tokens within multilingual sentences.
This strategy serves to bridge the semantic and syntactic gaps across different languages simultaneously, thus enabling more effective cross-lingual NER. 

We reformulate the \task{} task as a token pair relation classification problem. Given the source sentence $X = \{x_1,\dots,x_N\}$ of $N$ tokens, the relations between token pairs in $X$ are extracted into a relation set/matrix $R$, which can be categorized into intra-entity $R_{in}$ and extra-entity $R_{extra}$ relations. 
For an entity $E=(x_s,\dots,x_e)$ spanning from the $s$-th token to the $e$-th token ($1\le s \le e \le N$), $R(x_s,x_e)$ denotes the start-end relation. 
Additionally, $R(x_{i},x_{i+1})$ ($ s \le i < e$) signifies the neighbor relation between consecutive tokens within the entity. 
Note that the entity with just one word has a start-end relation and no neighbor relation. 
Here, $R(\cdot,\cdot)$ denotes the relation between two tokens. 
The intra-entity $R_{in}$ encompasses both start-end relations ($R(x_s,x_e)$) and neighbor relations ($R(x_{i},x_{i+1})$ ($ s \le i < e$)) within an entity. 
Moreover, relations such as $R(x_{i}, x_{j})$ ($ s \le i < e \land j \notin [s, e]$) represent the extra-entity relations.
\begin{BigEquation}
\begin{align}
    P(\mathcal{R}|X) = \prod_{1 < i,j < N \land i \neq j}P(R(x_i,x_j)|x_i,x_j;\Theta_{ner})
    \label{problem_definition}
\end{align}
\end{BigEquation}where the relation set $\mathcal{R}$ contains relations between the token $x_{i}$ and $x_{j}$ in the sentence. $\Theta_{ner}$ is the NER model parameter. 

In the cross-lingual setting, we amalgamate both source and target languages within a single sentence. Subsequently, we employ contrastive learning techniques to narrow the disparity between the source relation $R(x_i,x_j)$ and the corresponding aligned relations $R(y_a,y_b)$ in the target language. Simultaneously, we refine the representations of $x_{i:j}$ and $y_{a:b}$ through semantic contrastive learning. This multi-view contrastive learning process can be conceptualized as:
\begin{BigEquation}
\begin{align}
    \min (\lvert  R(x_i,x_j) - R(y_a,y_b) \rvert  + \lvert  F(x_i,x_j) - F(y_a,y_b) \rvert)
    \label{objective}
\end{align}
\end{BigEquation}where $R(\cdot)$ earns relation representation, $F(\cdot)$ gets semantic representation, $\lvert  R(x_i,x_j)-R(y_a,y_b) \rvert$ is relation distance, $\lvert  F(x_i,x_j) - F(y_a,y_b) \rvert$ is representation distance.

\section{Cross-Lingual Multi-view Contrastive Learning}
As shown in the overall architecture of Figure \ref{Fig.frame_2}, we resolve the cross-lingual NER task by distinguishing the relation between tokens and tokens for better modeling of intra-entity and extra-entity. We design a semantic contrastive objective to cluster the representations with the same meaning in different languages and introduce token-to-token relation contrastive learning for gathering similar token-to-token relations within entities across different languages. 
Following the previous work \cite{unitrans,CROP_Yang_2022}, the initial model is jointly trained on labeled source data and its code-switch counterparts (the translated data is a special case of code-switched data). Further, the target raw data is labeled by the previous model to build the synthetic corpora for the subsequent iteration of optimizing via self-training.
\subsection{Semantic Contrastive Learning}
For zero-resource \task{} without target annotated data, effectively representing the same entity with the cross-lingual pre-trained model in different languages becomes crucial to enhance the performance of the \task{} model. To address this challenge, we enrich the original source sentences by randomly substituting source phrases with target translations to create code-switched sentences comprised of both source and target tokens. This approach provides assistance for our proposed model to align the cross-lingual context of different languages and improve its ability to recognize named entities. Contrastive learning can be naturally and effectively used to reduce the gap among different languages under the cross-lingual scenario with code-switched data. Based on the code-switched corpora, we propose source-codeswitch contrastive learning to minimize the distance between positive samples and maximize the distance between negative samples. 
\begin{CJK}{UTF8}{gbsn}
Constructing a code-switched sentence for \task{} involves mixing entity names in both source and target languages to ensure that the code-switched sentence is fluent and accurate in terms of contextual information. 
For example, a code-switched sentence comprised of English and Chinese entity names can be ``我要去北京看看 the Great Wall (I want to go to Beijing to see the Great Wall)''. In this sentence, ``北京 (Beijing)'' is the Chinese location entity while ``the Great Wall'' is the English translation of the Chinese token ``长城''.
The code-switched sentence containing both source and target entities facilitates entity recognition in \task{}.
\end{CJK}

Given a source sentence $X=\{x_1,\dots,x_N \}$ and its code-switched sentence $Y=\{y_1,\dots,y_M\}$ (the target translation is a special case of the code-switched sentence), we consider them a positive sample pair within a batch and the source sentence is paired with other negative samples. The model is encouraged to learn language-agnostic representations for capturing the same meaning across different languages. The cosine similarity is used to measure the distance between sentence pairs. We apply an extra contextual fusion layer to refine the features from the cross-lingual pre-trained model to enhance the contextual information within the sequence. Supposing $H=\{h_1,\dots,h_N \}$ denote the final features of each token in source sentence $X$ and $H^{'}=\{h_1^{'},\dots,h_N^{'}\}$ denote the features of translated data, we obtain the sentence semantic representation $H_{avg}=f(H)$ and $H'_{avg}=f(H')$, where $f(\cdot)$ denotes the average or pooling operation. 

We introduce the contrastive learning between source and code-switched sentences, which helps model to reduce the gap of different language entities with the same meaning as:
\begin{equation}
\begin{aligned}
\mathcal{L}_{sc}= -\log\frac{e^{\text{sim}(H_{avg},H_{avg}')/\tau }}{ {\textstyle \sum_{j=1}^{B}}e^{\text{sim}(H_{avg},H_{avg}^{j})/\tau } }
\end{aligned}
\label{fun.sentence-level-cl}
\end{equation}
where $\text{sim}(\cdot)$ is cosine similarity, $B$ is the number of source data, $\tau$ is the temperature. $H_{avg}$ and $H'_{avg}$ are corresponding representations of the source and code-switched sentence.
\subsection{Token-to-Token Relation Contrastive Learning}
We introduce the token-to-token relation of our cross-lingual NER method and then leverage token-pair relation contrastive learning to cluster the representation of different languages. Token-pair relation is the important feature in the NER task \cite{OneRel,bs}, which can be used to enhance the cross-lingual entity recognition by constraining the cross-lingual token-pair relation.

Given a sentence $X=\{x_1,\dots, x_N \}$, we obtain the hidden representation $H=\{h_1\dots,h_N \}$ after the encoder layers. 
The relation representation $\{r_{ij}|(i,j\in[1,N])\} \in \mathbb{R}^{N\times N\times d_h}$ of token pair $(x_i,x_j)$ is then obtained through the biaffine token-to-token relation layer. 
\begin{align}
\begin{split}
&\tilde{h_i}=\text{MLP}(h_i) , \tilde{h_{j}}=\text{MLP}(h_j) \\
&r_{ij}=\tilde{h_i}^{\top}W_1\tilde{h_{j}} + W_2(\tilde{h_i}\oplus \tilde{h_{j}}) +b
\label{fun.s_o_object}
\end{split}
\end{align}
where $W_1$, $W_2$ and $b$ denote the trainable parameters, $\oplus$ means concatenation. $\text{MLP}$ is the fully connected layer.

The source or code-switched tokens are fed into the conditional layer normalization \cite{soie_2021_cln} module to obtain the relation representation $r_{ij}^{'}$ as:
\begin{equation}
\begin{aligned}
r_{ij}^{'} = \gamma_{i} &\odot (\frac{h_{j}-\mu }{\sigma}) + \lambda_{i} \\
\gamma _{i} = W_\alpha h_i + b_\alpha &, \lambda_{i} = W_\beta h_i + b_\beta \\
\mu =\frac{1}{d_h}\sum_{k=1}^{d_h}h_{jk}&, \sigma = \sqrt{\frac{1}{d_h}\sum_{k=1}^{d_h}(h_{jk}-\mu)^2} 
\end{aligned}
\label{fun.s_o_object_1}
\end{equation}
where $\gamma_{i}$ and $\lambda_{i}$ are from $h_i$. $\mu$ and $\sigma$ are the mean and standard deviation taken across $h_{j}$ elements respectively. $h_{jk}$ is the $k$-th dimension of $h_j$. $\odot$ is the element-wise product. $d_h$ is hidden size of $h_i$. $W_{\alpha}$, $W_{\alpha}$, $b_{\alpha}$, $b_{\beta}$ are learned parameters.

Finally, we take the token-to-token relation representations $r_{ij}$ and $r'_{ij}$ into an MLP layer to obtain their projection representations $z_{ij}$ and $z'_{ij}$. 
The contrastive learning objective of token-to-token relation is defined as:
\begin{equation}
\begin{aligned}
\mathcal{L}_{tc}=-\log\frac{e^{\text{sim}(z_{ij},z_{ij}^{'})/\tau }}{ {{\textstyle \sum_{j=1}^{2N}}e^{\text{sim}(z_{ij},z_{ij}^{'})/\tau }}}
\end{aligned}
\label{fun.t_cl-1}
\end{equation}
where contrastive learning applies to both source and target data, $\text{sim}(z_{ij}, z_{ij}')$ is the cosine similarity of source sentence and its counterparts (e.g. code-switched or source sentence).
\begin{table*}[ht]
\begin{adjustbox}{width=1\linewidth,center}
\fontsize{44pt}{32pt}\selectfont
\begin{tabular}{lcccccccccccccccccccc}
\toprule
\multicolumn{21}{c}{\textbf{mBERT}}\\ \midrule
\textbf{Method}  & \textbf{af} & \textbf{ar} & \textbf{bg} & \textbf{bn} & \textbf{de} & \textbf{el} & \textbf{es} & \textbf{et} & \textbf{eu} & \textbf{fa}   & \textbf{fi} & \textbf{fr} & \textbf{he} & \textbf{hi} & \textbf{hu} & \textbf{id} & \textbf{it} & \textbf{ja} & \textbf{jv} & \textbf{ka}     \\ \hline
mBERT \cite{bert}           & 76.9        & 44.5        & 77.1        & 68.8        & 78.8        & 71.6        & 74.0        & 76.3        & 68.0        & 48.2          & 77.2        & 79.7        & 56.5        & 66.9        & 76.0        & 46.3        & \underline{81.1}        & 28.9        & \underline{66.4}        & 67.7            \\
+Translate Train \cite{CROP_Yang_2022} & 74.5        & 37.6        & 77.8        & 73.2        & 77.2        & 74.9        & 69.4        & 74.1        & 63.2        & 43.1          & 75.9        & 76.1        & 55.4        & 68.1        & 77.2        & \underline{48.2}        & 77.2        & 36.6        & 55.1        & 64.4            \\
UniTrans \cite{unitrans}        & 78.2        & 47.0        & 79.5        & 74.6        & 79.8        & 75.6        & 75.2        & 76.5        & 67.2        & 49.3          & 75.6        & 80.1        & 58.4        & 72.1        & 77.9        & 44.6        & 78.3        & 37.6        & 56.2        & 69.9            \\
CROP  \cite{CROP_Yang_2022}           &\underline{81.0}        & \underline{48.0}        & \underline{80.8}        & \underline{74.9}        & \underline{80.3}        & \underline{78.7}        & \textbf{84.2}        & \underline{78.3}        & \underline{70.6}       & \textbf{63.2}          & \underline{79.1}        & \textbf{83.5}        & \textbf{64.7}        & \underline{77.1}        & \textbf{82.5}        & 46.4        & 79.9        & \underline{45.3}        & 57.7        & \textbf{74.1}            \\
\textbf{Ours}   & \textbf{82.2} 	& \textbf{50.3} 	& \textbf{81.1} 	& \textbf{79.0} 	& \textbf{82.3} 	& \textbf{78.9} 	& \underline{83.2} 	& \textbf{79.3} 	& \textbf{72.2} 	& \underline{54.9} 	& \textbf{79.4} 	& \underline{82.3} 	& \underline{61.9} 	& \textbf{78.9} 	& \underline{78.1} 	& \textbf{63.3} 	& \textbf{82.3} 	& \textbf{57.2} 	& \textbf{72.2} 	& \underline{70.4} \\ \hline
\textbf{Method}  & \textbf{kk} & \textbf{ko} & \textbf{ml} & \textbf{mr} & \textbf{ms} & \textbf{my} & \textbf{nl} & \textbf{pt} & \textbf{ru} & \textbf{sw}   & \textbf{ta} & \textbf{te} & \textbf{th} & \textbf{tl} & \textbf{tr} & \textbf{ur} & \textbf{vi} & \textbf{yo} & \textbf{zh} & \cellcolor{gray!20} \textbf{Avg$_{all}$} \\ \hline
mBERT \cite{bert}            & 50.4        & 60.2        & 53.7        & 56.2        & 61.9        & 47.6        & 82.1        & 79.6        & 65.2        & 72.8          & 50.8        & 46.8        & 0.4         & 71.2        & 75.5        & 36.9        & 69.7        & 51.7        & 44.1        &  \cellcolor{gray!20} 61.7            \\
+Translate Train \cite{CROP_Yang_2022}  & 48.2        & 61.2        & 61.0        & 58.7        & 67.5        & 57.3        & 79.6        & 78.4        & 61.2        & \underline{69.2} & 62.7        & 51.2        & 2.4         & 72.7        & 72.6        & 58.9        & 69.5        & 51.1        & 45.3        &  \cellcolor{gray!20} 62.3            \\
UniTrans \cite{unitrans}         & 52.5        & 61.4        & \underline{63.5}        & 62.3        & 65.8        & 59.2        & 82.4        & 80.3        & 64.8        & 65.2          & \underline{63.2}        & 56.1        & 3.1         & 73.4        & \underline{77.9}        & 64.1        & 69.7        & 50.1        & 47.4        &  \cellcolor{gray!20} 64.5            \\
CROP \cite{CROP_Yang_2022}            & \underline{54.9}        & \underline{62.6}        & \textbf{72.7}        & \textbf{70.6}        & \underline{71.1}        & \underline{61.3}        & \underline{84.6}        & \underline{81.7}        & \underline{69.7}        & 68.3          & \textbf{64.9}        & \textbf{61.6}        & \underline{3.9}         & \underline{76.9}        & \textbf{80.4}        & \textbf{78.0}        & \underline{70.0}        & \underline{51.8}        & \underline{54.4}        &  \cellcolor{gray!20} \underline{68.5}            \\
\textbf{Ours}     & \textbf{56.0} 	& \textbf{63.4} 	& 62.4 	& \underline{67.9} 	& \textbf{76.2} 	& \textbf{64.3} 	& \textbf{85.6} 	& \textbf{83.3} 	& \textbf{71.1} 	& \textbf{75.3} 	& 57.9 	& \underline{60.4}  	& \textbf{11.2} 	& \textbf{83.5} 	& 74.3 	& \underline{72.4} 	& \textbf{81.4} 	& \textbf{65.4} 	& \textbf{63.5} 	&  \cellcolor{gray!20} \textbf{70.4}                 \\ \bottomrule
\end{tabular}
\normalsize 
\end{adjustbox}
\caption{Experimental results on XTREME-40 initialized by pretrained cross lingual language model mBERT$_{base}$.}
\label{tab:xtreme_mbert_main_result}
\end{table*}
\begin{table*}[ht]
\begin{adjustbox}{width=1\linewidth,center}
\fontsize{44pt}{32pt}\selectfont
\begin{tabular}{lcccccccccccccccccccc}
\toprule
\multicolumn{21}{c}{\textbf{XLM-R} } \\ \midrule
\textbf{Method}  & \textbf{af}   & \textbf{ar}   & \textbf{bg}   & \textbf{bn}   & \textbf{de}   & \textbf{el}   & \textbf{es}   & \textbf{et}   & \textbf{eu}   & \textbf{fa}   & \textbf{fi}   & \textbf{fr}   & \textbf{he}   & \textbf{hi}   & \textbf{hu}   & \textbf{id}   & \textbf{it}   & \textbf{ja}   & \textbf{jv}   & \textbf{ka}     \\ \hline
XLM-R \cite{xlmr}           & 74.6          & 46.0          & 78.0          & 68.3          & 75.2          & 75.7          & 70.2          & 72.2          & 59.9          & 52.0          & 75.8          & 76.6          & 52.4          & 69.6          & 78.2          & 47.4          & 77.7          & 21.0          & 61.8          & 66.5            \\
+Translate Train \cite{CROP_Yang_2022}& 76.2          & 47.8          & 79.2          & 74.3          & 75.8          & 67.7          & 68.4          & 75.8          & 61.2          & 41.0          & 76.8          & 76.4          & 55.0          & 71.9          & 76.0          & \underline{50.6}          & 78.1          & 35.4          & 54.7          & 68.4            \\
UniTrans \cite{unitrans}          & 78.1          & \underline{48.1}          & 79.3          & 74.6          & 75.2          & 74.9          & 73.8          & 76.9          & 62.7          & 49.2          & 74.6          & 76.5          & 53.4          & 70.4          & 76.9          & 48.6          & 77.3          & 21.6          & \underline{62.2}          & 66.8            \\
CROP  \cite{CROP_Yang_2022}             & \textbf{80.3}          & 45.2          & \underline{80.4}          & \underline{75.7}          & \underline{79.6}          & \underline{78.5}          & \textbf{83.1}          & \underline{77.2}          & \underline{66.8}          & \textbf{65.5}          & \underline{77.9}          & \underline{82.9}          & \underline{63.5}          & \underline{77.4}          & \textbf{81.6}          & 46.1          & \underline{78.8}          & \underline{45.4}          & \textbf{63.2}          & \textbf{74.0}            \\
\textbf{Ours}    & \underline{79.7}          & \textbf{57.0}            & \textbf{81.5}          & \textbf{79.5}          & \textbf{80.2}          & \textbf{79.1}          &  \underline{79.9}             & \textbf{77.7}         & \textbf{67.1}          & \underline{55.8} &  \textbf{78.1}             & \textbf{83.0}       & \textbf{64.8}          & \textbf{78.1}          & \underline{78.5}              & \textbf{51.9}          & \textbf{79.9}              & \textbf{52.9}          & 61.5          & \underline{71.4}            \\ \hline
\textbf{Method}  & \textbf{kk}   & \textbf{ko}   & \textbf{ml}   & \textbf{mr}   & \textbf{ms}   & \textbf{my}   & \textbf{nl}   & \textbf{pt}   & \textbf{ru}   & \textbf{sw}   & \textbf{ta}   & \textbf{te}   & \textbf{th}   & \textbf{tl}   & \textbf{tr}   & \textbf{ur}   & \textbf{vi}   & \textbf{yo}   & \textbf{zh}   &  \cellcolor{gray!20} \textbf{Avg$_{all}$} \\ \hline
XLM-R \cite{xlmr}    & 43.2          & 49.9          & 62.3          & 59.6          & 67.3          & 53.5          & 80.2          & 78.1          & 64.3          & \underline{70.3}          & 55.0          & 50.1          & 3.0           & 69.4          & \underline{78.1}          & 63.6          & 68.2          & 47.5          & 27.7          &  \cellcolor{gray!20} 61.3            \\
+Translate Train \cite{CROP_Yang_2022} & 40.1          & 55.5          & 60.0          & 59.8          & 69.8          & 61.6          & 79.6          & 76.4          & 60.9          & 70.0          & 63.7          & 50.7          & 3.4           & 74.7          & 72.3          & 62.7          & 69.6          & 46.8          & 41.2          &  \cellcolor{gray!20} 62.3            \\
UniTrans \cite{unitrans}          & 46.5          & \underline{57.2}          & 65.5          & 64.5          & 70.2          & 62.6          & 81.8          & 79.4          & \underline{68.8}          & 68.9          & \underline{65.1}          & 56.1          & \underline{4.8}           & 74.8          & 76.4          & 71.0          & 69.8          & \underline{55.1}          & 44.4          &  \cellcolor{gray!20} 64.2            \\
CROP  \cite{CROP_Yang_2022}             & \underline{50.2}          & \textbf{59.8}          & \textbf{73.8}          & \textbf{71.6}          & \underline{71.8}          & \textbf{69.0}          & \textbf{83.5}          & \textbf{82.3}          & \textbf{70.2}          & 69.0          & \textbf{65.6}          & \underline{59.9}          & 3.1           & \underline{75.5}          &  \textbf{80.5}          & \underline{80.4}          & \underline{70.1}          & 52.6          & \underline{50.3}          &  \cellcolor{gray!20} \underline{68.2}            \\ 
\textbf{Ours}             & \textbf{53.2}          & 57.0          & \underline{69.1}          & \underline{67.1}          &\textbf{74.4}           & \underline{66.4}          & \underline{83.0}          &\underline{81.2}          & 65.1          & \textbf{71.8}          & 61.8         & \textbf{61.2}          & \textbf{19.2}           & \textbf{79.3}          & 76.2          &  \textbf{80.5}          & \textbf{76.7}          & \textbf{77.2}          & \textbf{60.3}          &  \cellcolor{gray!20} \textbf{69.7}           \\\bottomrule
\end{tabular}
\normalsize 
\end{adjustbox}
\caption{Experimental results on XTREME-40 initialized by pretrained cross lingual language model XLM-R$_{base}$.}
\label{tab:xtreme_xlmr_main_result}
\end{table*}
\subsection{Self-Training}
Due to the scarcity of annotated target language data, we use self-training to get pseudo labels for target data. 
First, our cross-lingual NER model $\Theta _{ner}^{src}$ is trained on the source labeled dataset $D_{x,r}^{src}=\{((x_i,x_j),r_{ij})\}_{i=1,j=1}^{N}$:
\begin{equation}
\begin{aligned}
\mathcal{L}_{ce}=-\frac{1}{N^{2}}\sum_i\sum_j\sum_{r=1}^{R}\hat{r_{ij}}\log P(r_{ij}=r|(x_{i},x_{j}))
\end{aligned}
\label{fun.ce-1}
\end{equation}
where $\hat{r_{ij}}$ is the gold relation label of the token pair $(x_i,x_j)$, $P(r_{ij}|(x_{i},x_{j})$ is the relation prediction probability of the token pair $(x_i,x_j)$. $R$ denotes the total number of relation classes.
The final loss can be formed of multiple contrastive learning losses and the token-to-token relation class loss. The final loss of the CrossNER model is accumulated as:
\begin{equation}
\begin{aligned}
\mathcal{L}_{\Theta _{ner}^{src}}=\mathcal{L}_{ce}^{src}+w(\mathcal{L}_{sc}^{src}+\mathcal{L}_{tc}^{src})
\end{aligned}
\label{fun.L_tea}
\end{equation}
Then, the model $\Theta_{ner}^{src}$ generates pseudo labels of token pairs for unlabeled target dataset $D_{x}^{tgt}=\{(x_i)\}_{i=1}^{M} $. 
The model is trained on the source and target pseudo dataset. Since pseudo-labels of the target data may bring extra noise, we only retain the contrastive learning of token pair relation and minimize the mean squared error \cite{ren22_mse} of the source and target model on the prediction distribution. The training objective of the model $\Theta_{ner}^{tgt}$ is defined as:
\begin{equation}
    \begin{aligned}                     \mathcal{L}_{tgt}=\mathcal{L}_{ce}^{tgt}+w_{1}\mathcal{L}_{tc}^{tgt}+w_{2}\mathcal{L}_{mse}  \end{aligned}
\label{fun.L_tea1}        
\end{equation}
\begin{equation}
    \begin{aligned}             
        \mathcal{L}_{mse}=\frac{1}{N^{2}}\sum_{i=1}^{N} \sum_{j=1}^{N} | P(r_{ij}|\Theta _{ner}^{src}),P(r_{ij}|\Theta _{ner}^{tgt})| 
    \end{aligned}
\label{fun.L_tea2}        
\end{equation}
where the loss $\mathcal{L}_{ce}^{tgt}$ of the target model is similar to Equation \ref{fun.ce-1}, $\mathcal{L}_{tc}^{tgt}$ likes Equation \ref{fun.t_cl-1}, $P(r_{ij}|\Theta)$ is the predicted probability of the token pair relation under the model $\Theta$, $w_{1}$ and $w_{2}$ are the hyperparameters. $|\cdot|$ is the MSE distance.
\section{Experiments}
\subsection{Datasets}
\subsubsection{XTREME-40} 
The proposed method is evaluated on the XTREME benchmark \cite{hu2020xtreme}.
Following the previous work \cite{hu2020xtreme}, we use the same split for the train, validation, and test sets, including the \texttt{LOC}, \texttt{PER}, and \texttt{ORG} tags. 
All NER models make the English training data the source language and evaluate other languages data.
\subsubsection{CoNLL} We also run experiments on CoNLL-02 and CoNLL-03 datasets \cite{conll02,conll03} covering 4 languages: Spanish (es), Dutch (nl), English (en), and  German (de).  
The datasets have \texttt{LOC}, \texttt{ORG}, \texttt{MISC}, and \texttt{PER} entity types. 
We split them into the train, validation, and test sets, following the prior work \cite{CROP_Yang_2022}.
\subsection{Implementation Details and Evaluation}
For a fair comparison, we adopt the same structure and model size, which all have 12 layers with an embedding dimension of 768 under the base architecture of both mBERT \cite{bert} and XLM-R \cite{xlmr}. We set the batch size as 32 for XTREME-40 and CoNLL. We use AdamW \cite{adamw} for optimization with a learning rate of $1e^{-5}$ for the pre-trained model and $1e^{-3}$ for other extra components. The dimension of the projection representations for contrastive learning is set to 128. 
We use average entity-level valid F1 scores of all languages to choose the best checkpoint and report the F1 scores on all test sets. 
We compare our approach with the different strong baselines UniTrans \cite{unitrans}, CROP \cite{CROP_Yang_2022}, Translate-Train \cite{CROP_Yang_2022}, and TSL \cite{Multi_source_2020}, which are initialized by the cross-lingual pre-trained model mBERT and XLM-R. 
We set a threshold and remove samples below it to mitigate the noise raised by pseudo-label data. 
Besides, we eliminate data that only has the ``O'' label and use the continuity of the inner relation of the entity to remove some discontinuous entity data.

\subsection{Main Results}
\subsubsection{XTREME-40} We present the results on the XTREME-40 dataset in Table \ref{tab:xtreme_mbert_main_result} and \ref{tab:xtreme_xlmr_main_result} by different cross-lingual pre-trained language models including mBERT and XLM-R. Overall, the average F1 score of our method outperforms the previous baselines by a large margin. 
Compared with the methods UniTrans and CROP initialized by mBERT, our work significantly improves +1.9 F1 points. 
For languages id (Indonesian), zh (Chinese), and ja (Japanese) distant from English, our method can further gain +10 points improvement than CROP. 
It is due to the effectiveness of the shared representations across the different languages by multi-view contrastive learning. 
For the methods initialized by the XLM-R, our model also gets a consistent promotion by +1.5 points compared to the baseline CROP in the average F1 score. 

\subsubsection{CoNLL} Table \ref{tab:conll_main_result} shows the experimental results on CoNLL dataset. To make a fair assessment, we use the mBERT base model. 
Compared with TSL, our approach achieves an average F1 score improvement of +2.3.
An averaged +1.3 F1 improvement is gained compared to UniTrans and CROP, which means the effectiveness of our proposed multi-view contrastive learning comprised of semantic and token pair relation contrastive. 
In contrast to the previous models, our method demonstrates favorable enhancements by reducing the gap among different languages in the shared space and clustering cross-lingual entities with the same meaning.
\begin{table}[t]
\begin{adjustbox}{width=1\columnwidth,center}
\fontsize{28pt}{24pt}\selectfont
\begin{tabular}{lccc>{\columncolor{gray!20}}c}
\toprule
\textbf{Method}            & \textbf{de} & \textbf{es} & \textbf{nl} & \textbf{Avg} \\ \hline
X-Lingual Clusters (T{\"{a}}ckstr{\"{o}}m et al. 2012)     & 40.4        & 59.3        & 58.4        & 52.7             \\
+Wikifier (Tsai et al. 2016)         & 48.1        & 60.6        & 61.6        & 56.8             \\
Inverted Softmax \cite{Smith2017}         & 58.5        & 65.1        & 65.4        & 63.0             \\
Cheap Translation (Mayhew et al. 2017)       & 57.2        & 64.1        & 63.4        & 61.6             \\
BWET \cite{Xie_2018}          & 57.8        & 72.4        & 71.3        & 67.2             \\
TMP (Jain et al. 2019)          & 65.2        & 75.9        & 74.6        & 71.9             \\
mBERT \cite{bert}          & 75.0        & 74.6        & 77.9        & 75.8             \\
BERT-f \cite{wu-Dredze}         & 71.1        & 74.5        & 79.5        & 75.0             \\
XLM-R \cite{xlmr}          & 73.4        & 77.4        & 78.9        & 76.6             \\
Cross-Augmented (Bari et al. 2020b)            & 61.5        & 73.5        & 69.9        & 68.3             \\
Meta-learning-based \cite{wu2020enhanced}    & 73.2        & 76.8        & 80.4        & 76.8             \\ 
TSL \cite{Multi_source_2020} & 75.3 	     & 78.0 	      & 81.3 	     & 78.2      \\ 
UniTrans \cite{unitrans}   & 74.8        & \textbf{79.3}        & \textbf{82.9}        & 79.0             \\
MulDA \cite{mulda}         & 78.2        & 77.5        & 78.4        & 78.0             \\
+Translate Train \cite{CROP_Yang_2022}          & 74.2        & 77.8        & 79.2        & 77.1             \\
CROP \cite{CROP_Yang_2022} & \underline{80.1}        & 78.1        & 79.5        & \underline{79.2}             \\  \hline
\textbf{Ours}              & \textbf{81.0}        & \underline{79.2}        & \underline{81.3}        & \textbf{80.5}             \\ \bottomrule
\end{tabular}
\normalsize 
\end{adjustbox}
\caption{Experimental results on CoNLL.}
\label{tab:conll_main_result}
\end{table}
\begin{figure}[t]
\begin{center}
    \includegraphics[width=1\columnwidth]{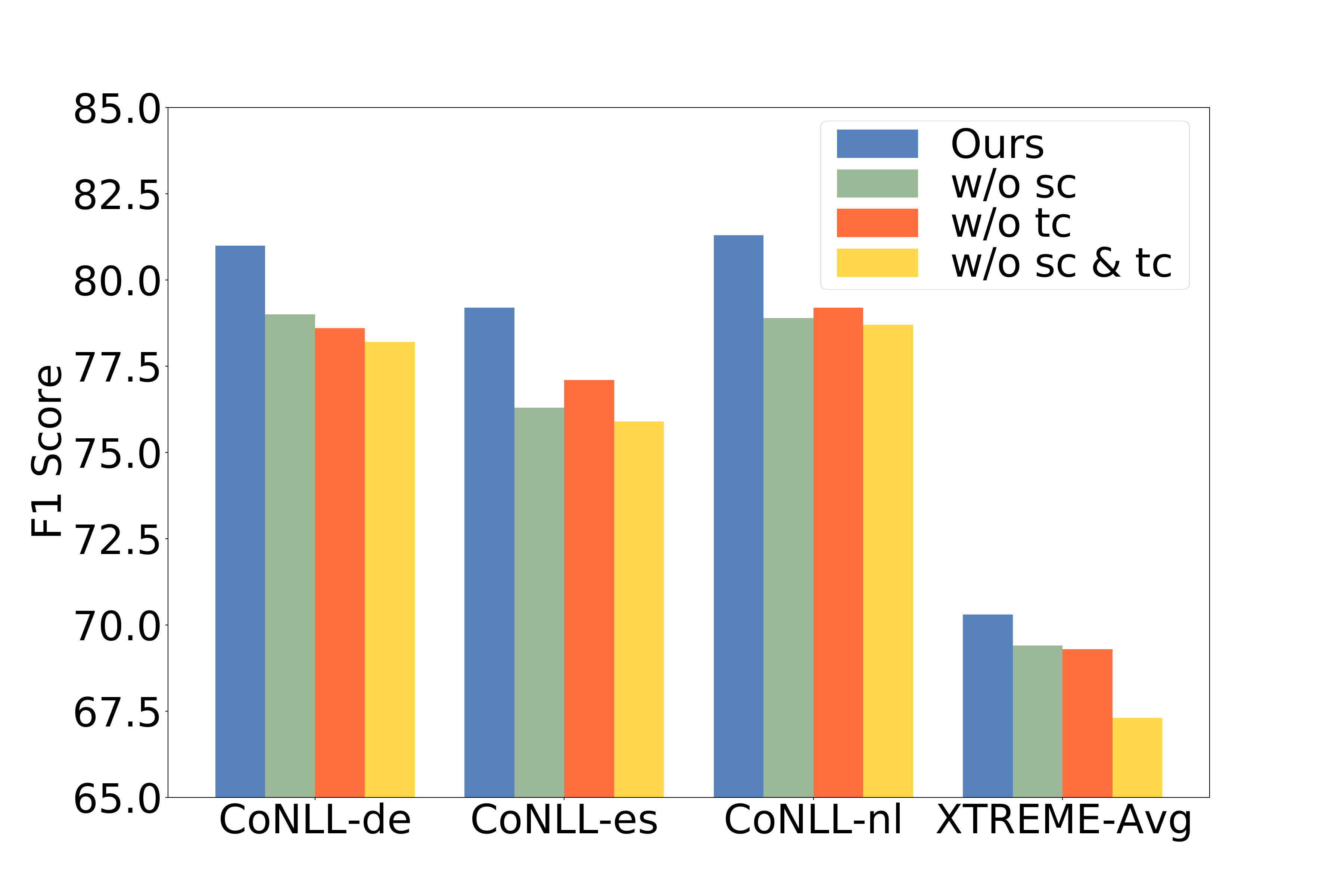}
    \caption{Ablation study of \ourmethod{}. XTREME-Avg denotes the average F1 scores of 39 languages in XTREME. }  
    \label{Fig.ablation_study} 
\end{center}
\end{figure}
\section{Analysis}
\subsubsection{Ablation Study}
To verify the effectiveness of \ourmethod{}, we introduce the following series of ablation experiments: {\large{\ding{172}}} Ours, which is the final model with the multi-view contrastive objectives;  {\large{\ding{173}}} w/o sc, which adopts token-to-token relation contrastive learning; {\large{\ding{174}}} w/o tc, which use the semantic contrastive objective; {\large{\ding{175}}} w/o sc \& tc, where semantic and token pair relation contrastive learning are removed. From the ablation experiments in Figure \ref{Fig.ablation_study}, our method outperforms {\large{\ding{173}}}, {\large{\ding{174}}}, and {\large{\ding{175}}}. {\large{\ding{173}}} w/o sc, which shows aligned semantic representations in different languages play a pivotal role in \task{} by enhancing shared semantic representations. {\large{\ding{174}}} w/o tc gets worse performance compared to our method {\large{\ding{172}}}, indicating that the contrastive learning of relation between tokens clusters the representations of the similar entities. {\large{\ding{175}}} w/o sc \& tc has the worst performance, which verifies that a multi-view of semantic and relation contrastive learning between tokens is the best strategy. 
\subsubsection{Representation on Token-to-Token Relationships} 
\begin{figure}[t]
    \centering
    \subfigure[]{
        \centering
        \includegraphics[width=0.47\columnwidth]{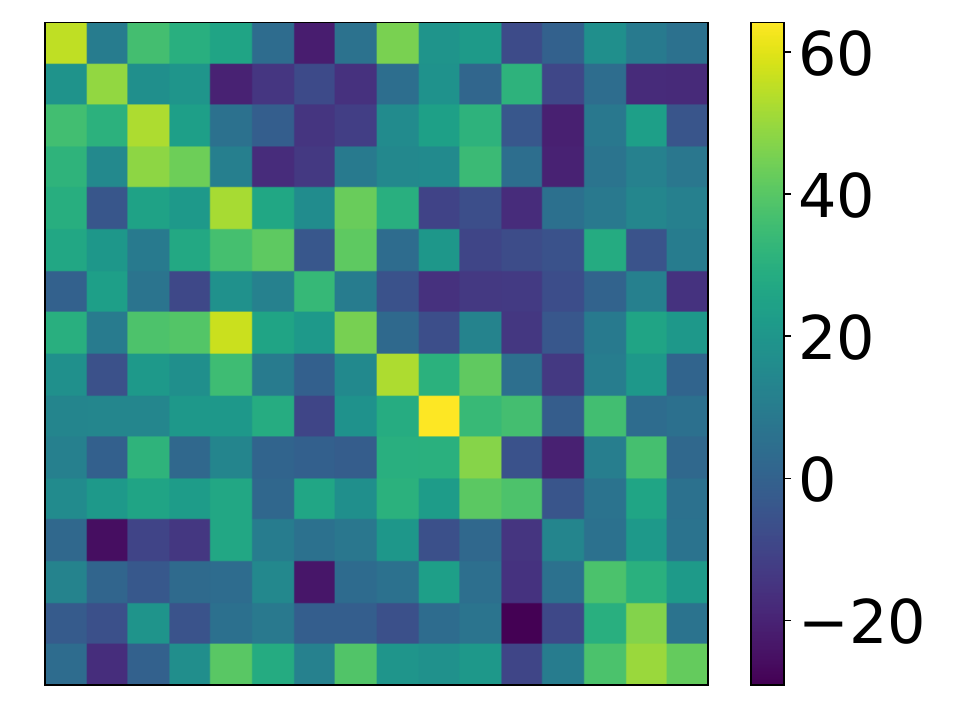}
        \label{Fig.cl_relation_1}
    }
    \subfigure[]{

        \centering
        \includegraphics[width=0.47\columnwidth]{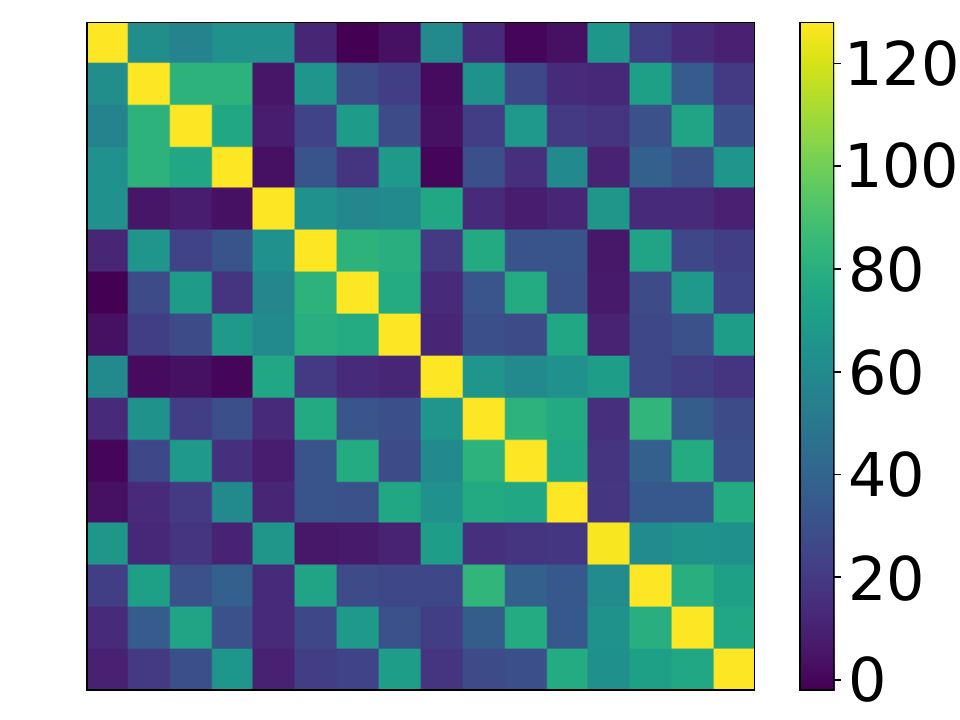}
        \label{Fig.cl_relation_2}    
    }
    \caption{\subref{Fig.cl_relation_1} denotes the representation on token-to-token relation without contrastive learning and \subref{Fig.cl_relation_2} is the representation on token-to-token relation with contrastive learning. }
    \label{Fig.cl_relation}
\end{figure}

To intuitively understand the effectiveness of token-to-token relation representation contrastive, we take a close look at the token-to-token relations within a sentence. We visualize the distance between the generated representations of the token-to-token relationships without contrastive learning and the counterparts with our method. The distance matric (cosine similarity) is shown in Figure \ref{Fig.cl_relation}. It can be seen that the similarity between the token-token relationship features of the baseline displays a random pattern. However, two slashes of Figure \ref{Fig.cl_relation_2} with higher brightness appear in the similarity matrix in a fixed pattern. The positive sample pair has a higher score with brightness, achieving the desired effect.
\begin{figure}[t]
    \centering
    \subfigure[]{
        \centering
        \includegraphics[width=0.47\columnwidth]{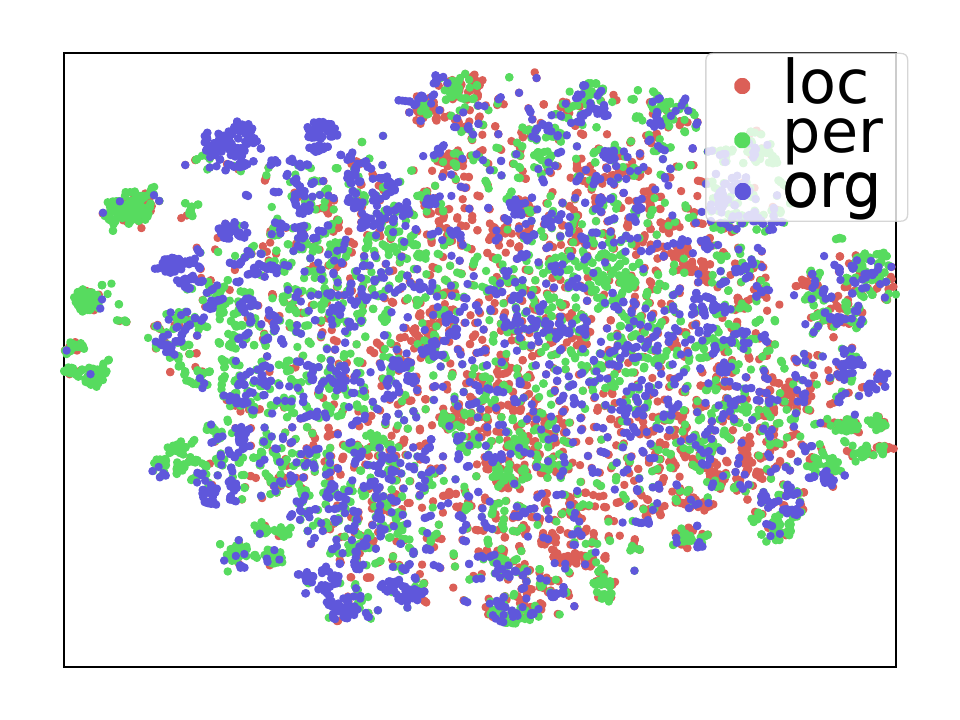}
        \label{Fig.zh_tsne_1}
    }
    \subfigure[]{
        \centering
        \includegraphics[width=0.47\columnwidth]{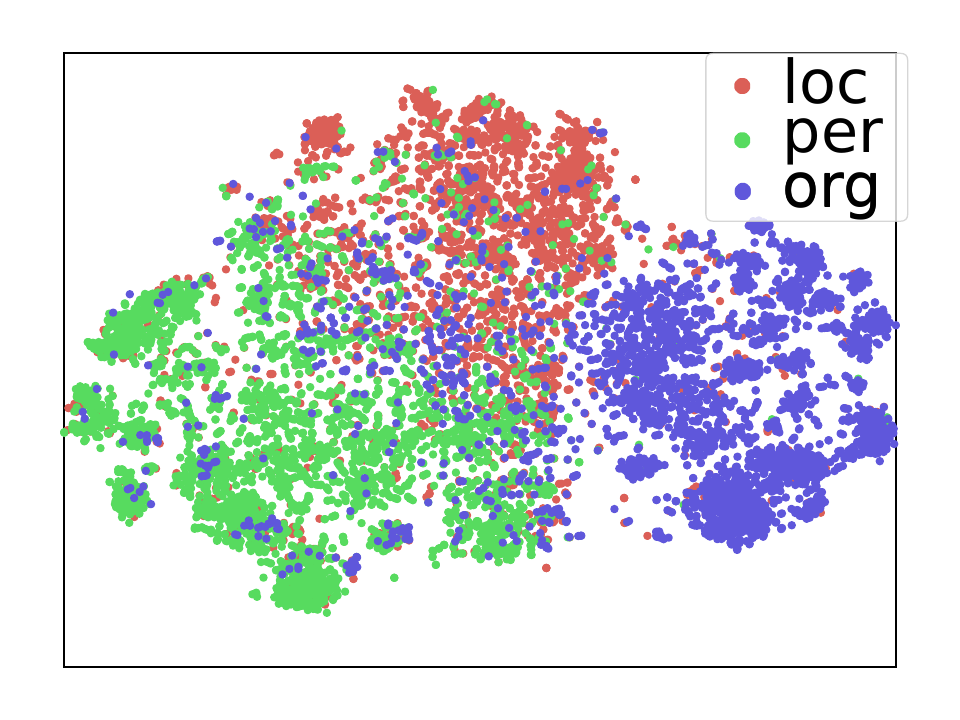}
        \label{Fig.zh_tsne_2}      
    }
    \caption{t-SNE visualization of different pre-defined categories (e.g. \texttt{LOC}) in Chinese. \subref{Fig.zh_tsne_1} and \subref{Fig.zh_tsne_2} indicate the token-to-token relation representations between entities w and w/o contrastive learning respectively.}
    \label{Fig.tsne} 
\end{figure}
\begin{figure*}[ht]
    \centering
    \subfigure[]{
    \includegraphics[width=0.66\columnwidth]{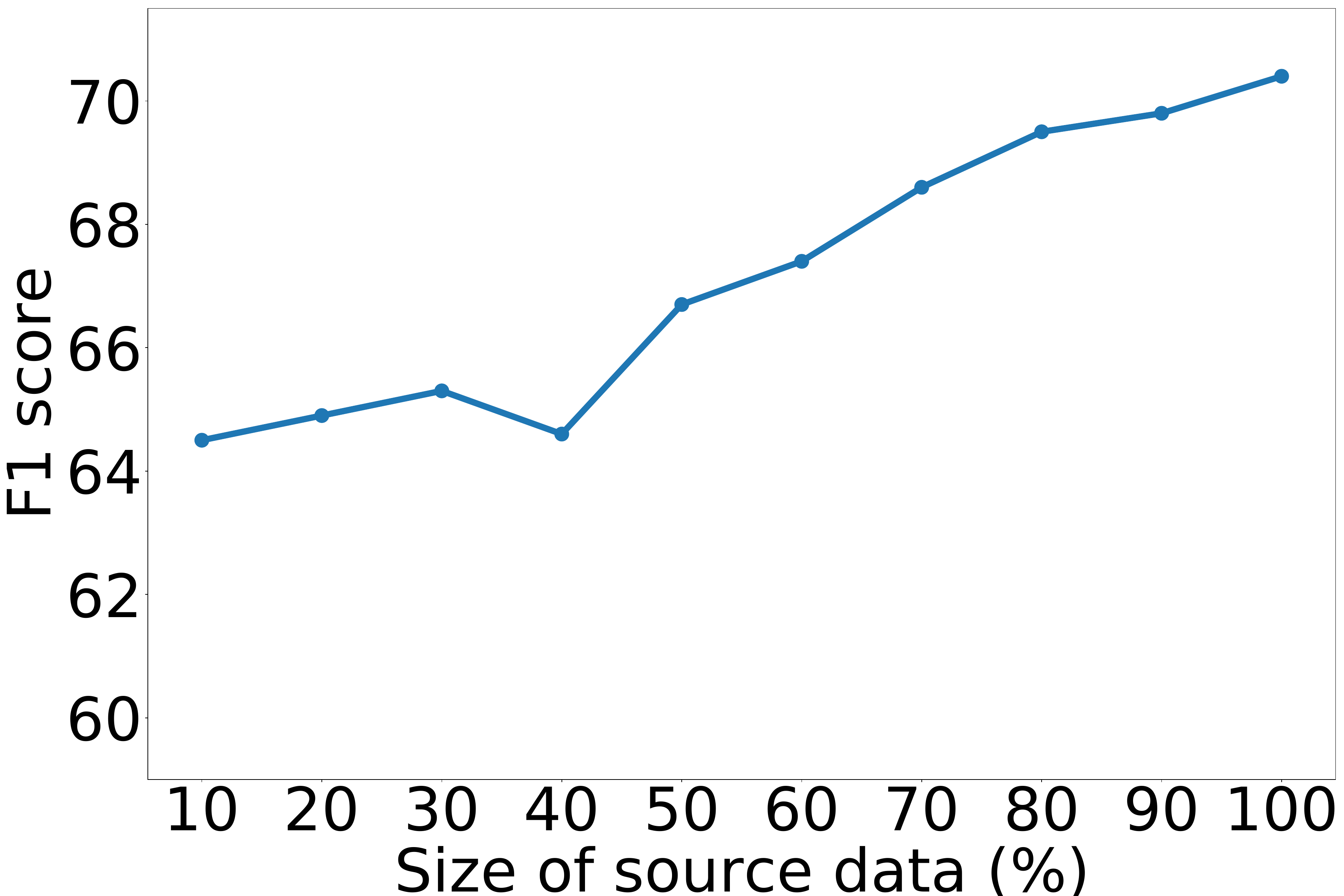}
    \label{Fig.f_trainsize_1}
    }
    \subfigure[]{
    \includegraphics[width=0.66\columnwidth]{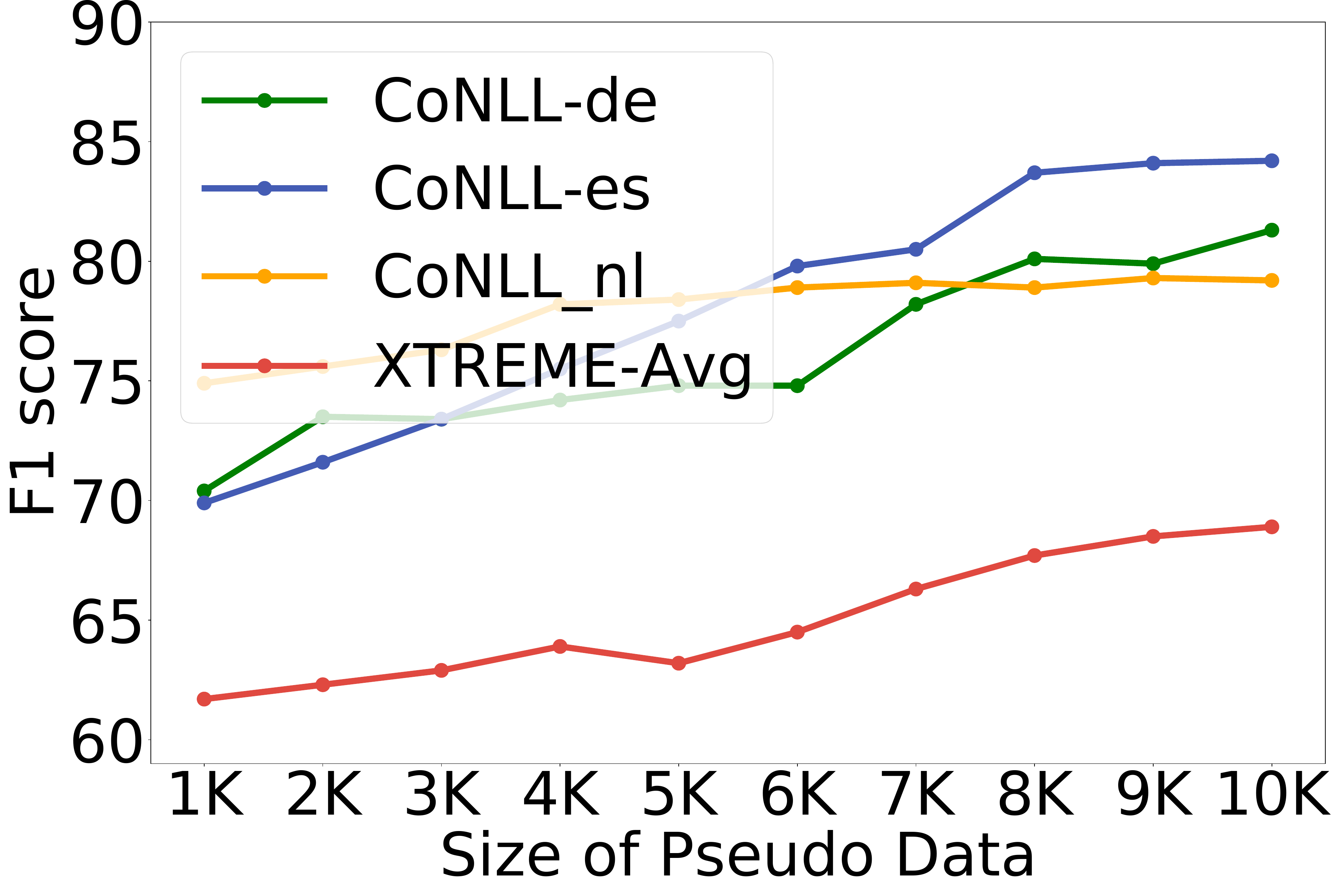}
    \label{Fig.f_trainsize_2}
    }
    \subfigure[]{
    \includegraphics[width=0.66\columnwidth]{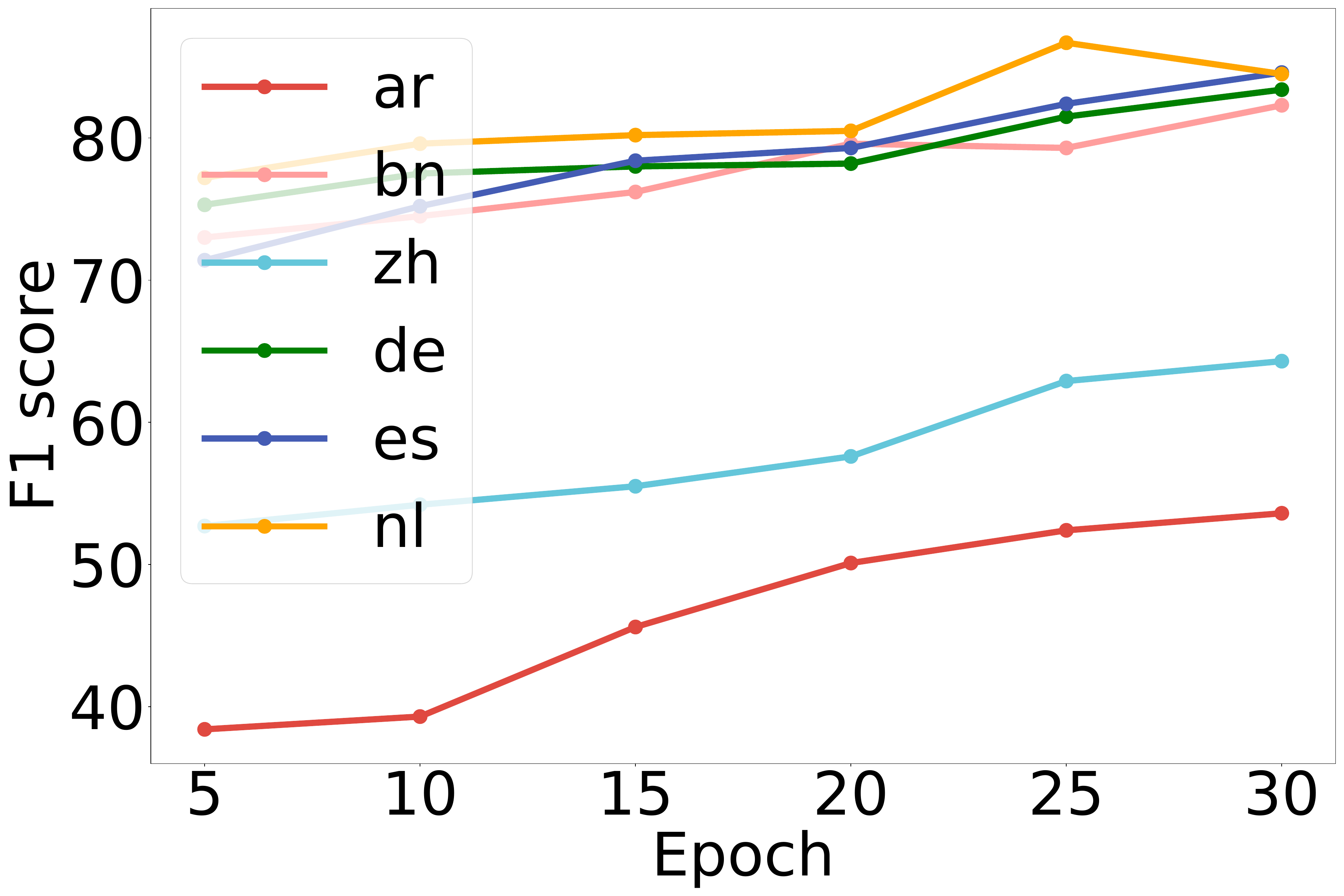} 
    \label{Fig.f_for_PseudoLabelQuality}
    }
    \caption{\subref{Fig.f_trainsize_1} Evaluation results on the validation sets with different source data training sizes. \subref{Fig.f_trainsize_2} Evaluation results on the validation sets with different pseudo-label target data training sizes. \subref{Fig.f_for_PseudoLabelQuality} Pseudo label quality for unlabeled target data. } 
\end{figure*}
\begin{figure*}[ht] 
    \centering
    \includegraphics[width=0.8\linewidth]{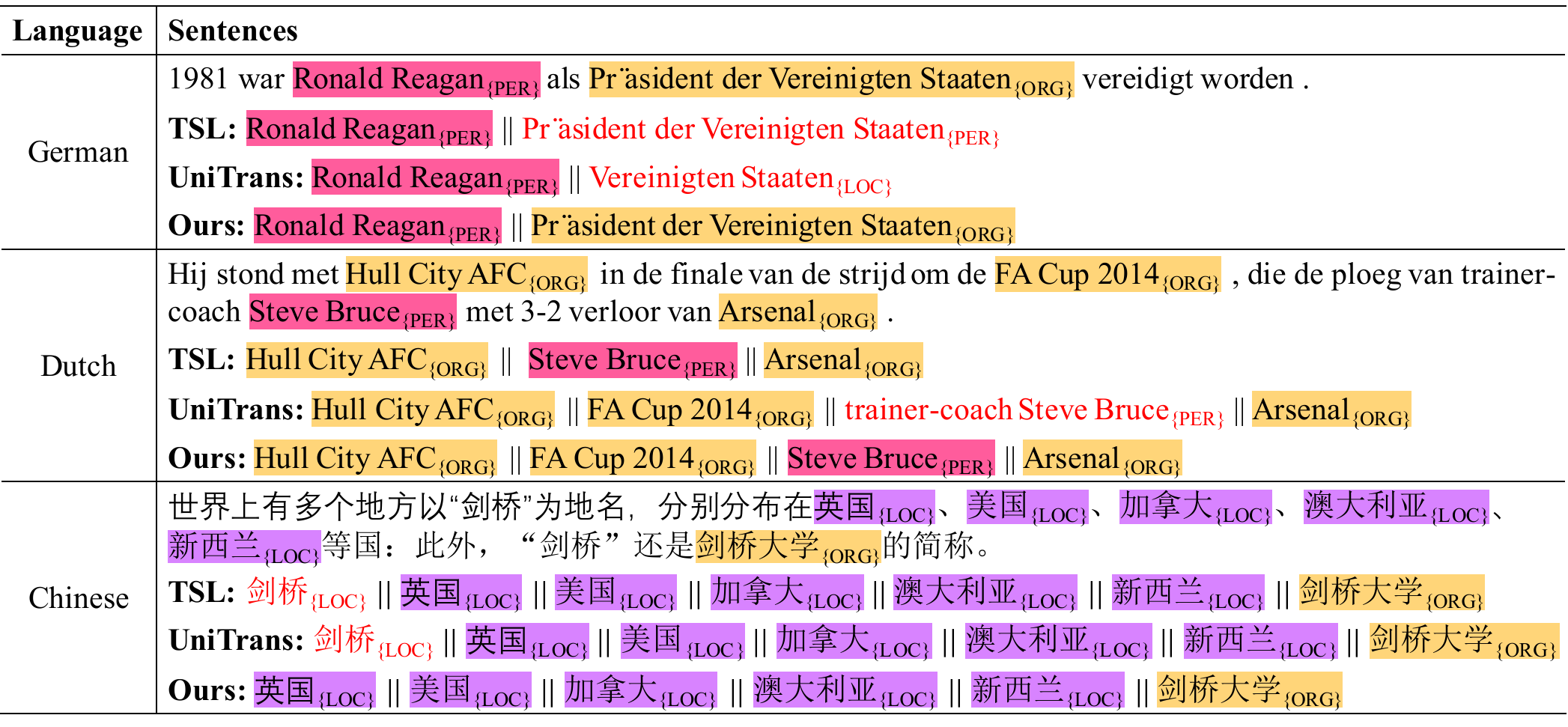} 
    \caption{Case study on \task{}. Texts with colorful backgrounds are gold entities, and the red fonts are incorrect predictions. }
    \label{Fig.case_study}
\end{figure*}
\subsubsection{Distribution of Multilingual Corpora}
We conduct a t-SNE visualization \cite{tsne} of token-to-token relation representations. 
Figure \ref{Fig.zh_tsne_1} implies that the model w/o multi-view contrastive learning generates token pair relation representations between different classes. 
Figure \ref{Fig.zh_tsne_2} shows that \ourmethod{} w/ multi-view contrastive objectives produce more distinct and distinguishable representations, where entities can be separated into independent regions for better performance of \task{}. 
\subsubsection{Effect of Training Data Size} We consider the impact from two aspects: the amount of source labeled data and the pseudo-label target data.  
Randomly sample to build the source data size ratio for analytic experiments according to a proportion from $10\%, 20\%,\dots,100\%$ in Figure \ref{Fig.f_trainsize_1}. In Fgiure \subref{Fig.f_trainsize_2}, we randomly sample the pseudo-label target data and put the train data size to $\{1K, 2K, \dots, ALL\}$ sentences to train the model. From Figure \ref{Fig.f_trainsize_1}, surprisingly, training with only 10\% of the source data brings huge gains. In Figure \ref{Fig.f_trainsize_2}, the overall model performance continues to be improved with increasing pseudo-label target data since the disturbance of pseudo-label data noise may cause slight fluctuations in training. When the pseudo-label target data size grows to a certain extent, the progress of the \task{} model becomes smaller. The reason is that the model can not learn more useful information from sufficient data unless given new valuable knowledge for \task{}. 
\subsubsection{Pseudo Label Quality}
The quality of pseudo-labels is crucial for the success of self-training cross-lingual NER. To evaluate the effectiveness of our method in improving pseudo-label quality, we use the gold labels of the unlabeled target language data as a reference to measure the F1 score after each epoch. We select representative languages from different language families, including ar, bn, zh, de, es, and nl from XTREME-40. 
In Figure \ref{Fig.f_for_PseudoLabelQuality}, Our experiments show that the quality of pseudo labels about target data is improved with the increasing training epoch, proving the effectiveness of our multi-view contrastive learning cross-lingual NER method. Especially a significant improvement in the F1 score of data zh can be observed, raising nearly 20 points. 
\subsubsection{Case Study}
\begin{CJK}{UTF8}{gbsn}
We select some cases of target languages that are similar and distant from the source language in Figure \ref{Fig.case_study}. In German, TSL and UniTrans have similar mistakes in the \texttt{ORG} entity ``Präsident der Vereinigten Staaten''. TSL incorrectly tags the entity type, while UniTrans mislabels it and decides the entity span wrongly. In Dutch, UniTrans has an error in the \texttt{PER} entity span, and TSL misses an \texttt{ORG} entity. Our work gains the right prediction by learning the relation of token pairs and achieving explicit recognition with token-to-token relation contrastive learning. For Chinese, TSL and UniTrans recognize the first ``剑桥'' as a \texttt{LOC} entity, while our model gets the right prediction. It can be attributed to our method getting better semantics than baselines.
\end{CJK}
\section{Conclusion}
In this work, we propose \ourmethod{}, a multi-view contrastive learning framework for the cross-lingual NER comprising semantic contrastive learning and token-to-token relation contrastive learning. 
Specifically, we construct the code-switched data by randomly substituting some phrases with the target counterparts for the semantic contrastive learning of the source and the corresponding code-switched sentence. 
Further, token-to-token contrastive learning enhances the syntactic representation of entities in different languages. 
Contrastive learning objectives minimize the semantic gap across different languages even for distant languages and improve the cross-lingual recognition performance with the more distinct entity representations. 
Extensive Experimental results show that our approach greatly performs better than baselines by a large margin on the XTREME-40 and CoNLL benchmarks.

\section{Ethical Statement}
For ethical considerations, our method, \ourmethod{}, comprises semantic and token-to-token relation contrastive learning for the cross-lingual NER. 
Experiments are conducted on public datasets from scientific papers with no identity characteristics.

\section{Acknowledgments}
This work was supported in part by the National Natural Science Foundation of China (Grant Nos. 62276017, U1636211, 61672081), and the Fund of the State Key Laboratory of Software Development Environment (Grant No. SKLSDE-2021ZX-18). Jingang Wang is funded by Beijing Nova Program(Grant NO. 20220484098).

\appendix
\bibliography{crossner}

\end{document}